\documentclass[sigconf]{acmart}

\usepackage{graphicx}
\usepackage[caption=false]{subfig}
\usepackage{tfrupee}
\usepackage{balance}

\renewcommand\footnotetextcopyrightpermission[1]{} 
\def\BibTeX{{\rm B\kern-.05em{\sc i\kern-.025em b}\kern-.08emT\kern-.1667em\lower.7ex\hbox{E}\kern-.125emX}}

\newif{\ifhidecomments}
\hidecommentsfalse
\ifhidecomments
      \newcommand{\amit}[1]{}
 \else
      \newcommand{\amit}[1]{\textcolor{blue}{[#1 ---\textsc{am}]}}
 \fi
 
\newcount\Comments
\newcommand{\bt}[1]{\ifnum\Comments=1{\color{purple}{[BT: #1]}}\fi}
\newcommand{\azc}[1]{\ifnum\Comments=1{\color{purple}{[AZC: #1]}}\fi}
    
\begin{document}

%
\title[Learning from Digital Adherence Data]{Learning to Prescribe Interventions for Tuberculosis Patients Using Digital Adherence Data}

%
\author{Jackson A. Killian}
\email{jakillia@usc.edu}
\orcid{0000-0001-8555-1327}
\affiliation{%
  \institution{University of Southern California}
  \streetaddress{941 Bloom Walk}
  \city{Los Angeles}
  \state{California}
  \country{USA}
  \postcode{90089}
}

\author{Bryan Wilder}
\email{bwilder@usc.edu}
\affiliation{%
  \institution{University of Southern California}
  \streetaddress{941 Bloom Walk}
  \city{Los Angeles}
  \state{California}
  \country{USA}
  \postcode{90089}
}

\author{Amit Sharma}
\email{amshar@microsoft.com}
\affiliation{%
  \institution{Microsoft Research India}
  \streetaddress{#9, Lavelle Road}
  \city{Bangalore}
  \state{Karnataka}
  \country{India}
  \postcode{}
}

\author{Daksha Shah}
\email{Cooldrdax@gmail.com}
\affiliation{%
  \institution{MCGM Mumbai}
  \streetaddress{}
  \city{Mumbai}
  \state{Maharashtra}
  \country{India}
  \postcode{}
}

\author{Vinod Choudhary}
\email{drvinod06@gmail.com}
\affiliation{%
  \institution{RNTCP, Mumbai}
  \streetaddress{}
  \city{Mumbai}
  \state{Maharashtra}
  \country{India}
  \postcode{}
}

\author{Bistra Dilkina}
\email{dilkina@usc.edu}
\orcid{0000-0002-6784-473X}
\affiliation{%
  \institution{University of Southern California}
  \streetaddress{941 Bloom Walk}
  \city{Los Angeles}
  \state{California}
  \country{USA}
  \postcode{90089}
}
\author{Milind Tambe}
\email{tambe@usc.edu}
\orcid{0000-0003-3296-3672}
\affiliation{%
  \institution{University of Southern California}
  \streetaddress{941 Bloom Walk}
  \city{Los Angeles}
  \state{California}
  \country{USA}
  \postcode{90089}
}

%
\renewcommand{\shortauthors}{Killian, et al.}

%
\begin{abstract}
Digital Adherence Technologies (DATs) are an increasingly popular method for verifying patient adherence to many medications. We analyze data from one city served by 99DOTS, a phone-call-based DAT deployed for Tuberculosis (TB) treatment in India where nearly 3 million people are afflicted with the disease each year. The data contains nearly 17,000 patients and 2.1M dose records. We lay the groundwork for learning from this real-world data, including a method for avoiding the effects of unobserved interventions in training data used for machine learning. We then construct a deep learning model, demonstrate its interpretability, and show how it can be adapted and trained in different clinical scenarios to better target and improve patient care. In the real-time risk prediction setting our model could be used to proactively intervene with 21\% more patients and before 76\% more missed doses than current heuristic baselines. For outcome prediction, our model performs 40\% better than baseline methods, allowing cities to target more resources to clinics with a heavier burden of patients at risk of failure. Finally, we present a case study demonstrating how our model can be trained in an end-to-end decision focused learning setting to achieve 15\% better solution quality in an example decision problem faced by health workers.
\end{abstract}

%
%


%

%
\maketitle

\section{Introduction}
\label{Introduction}
The World Health Organization (WHO) reports that tuberculosis (TB) is among the top ten causes of death worldwide \cite{GTR18}, yet in most cases it is a curable and preventable disease. The prevalence of TB is caused in part by non-adherence to medication, resulting in greater risk of death, reinfection and contraction of drug-resistant TB \cite{thomas2005predictors}. To combat non-adherence, the WHO recommends directly observed treatment (DOT), in which a health worker directly observes and confirms that a patient is consuming the required medication daily. However, requiring patients to travel to the DOT facility causes financial burden, and potentially social stigma due to public fear of the disease. Such barriers contribute to patients being lost to follow up, making TB eradication difficult. Thus, digital adherence technologies (DATs), which give patients flexible means to prove adherence, have gained global popularity \cite{subbaraman2018digital}.

\begin{figure}[t!]
  \includegraphics[width=0.4\textwidth]{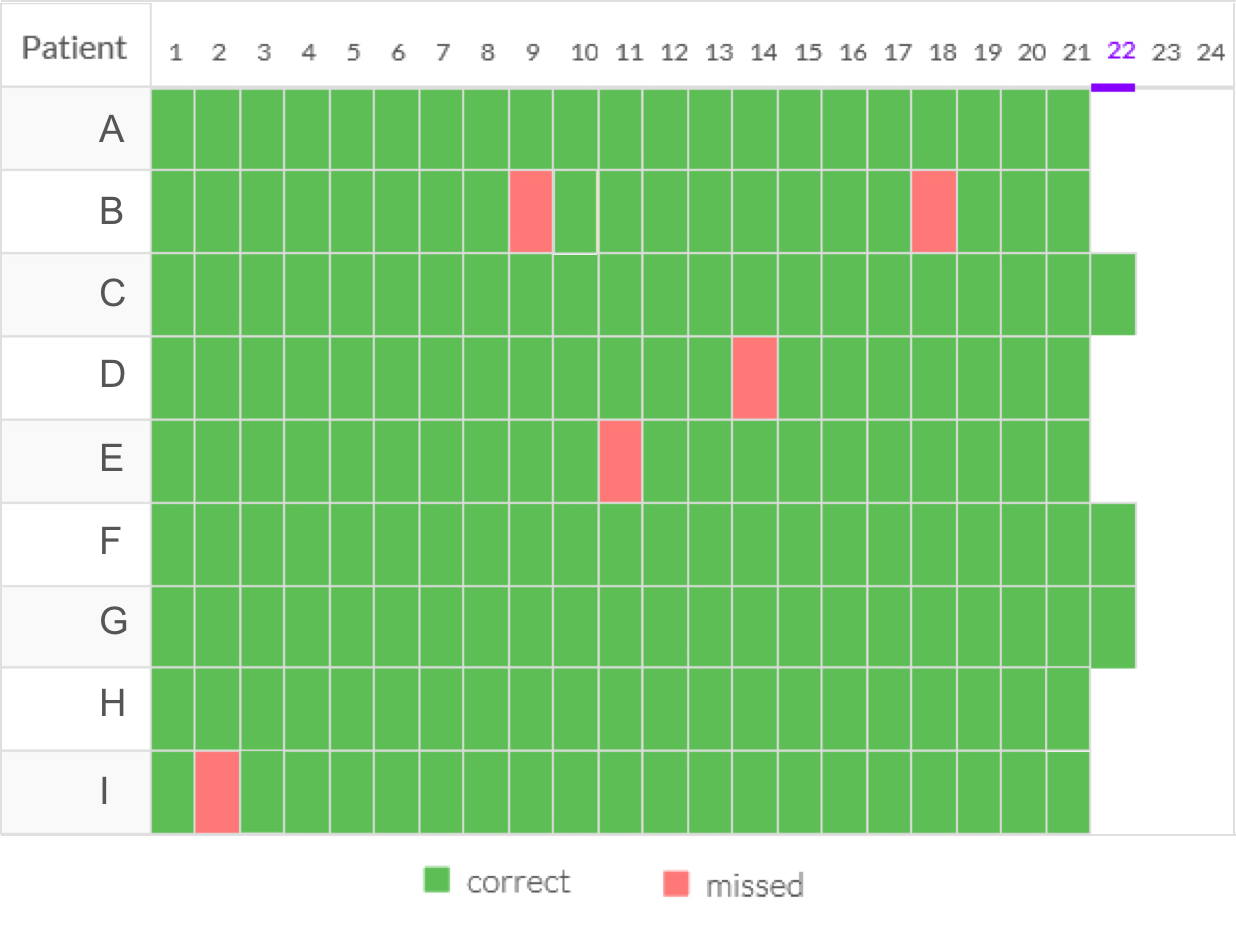}
  \caption{99DOTS electronic adherence dashboard seen by health workers for a given month. Missed doses are marked in red while consumed doses are marked in green.}
  \label{DotsDash}
\end{figure}

DATs allow patients to be "observed" consuming their medication electronically, e.g.\ via two-way text messaging, video capture, electronic pillboxes, or toll-free phone calls. Health workers can then view real-time patient adherence on a dashboard such as \textbf{Figure \ref{DotsDash}}. In addition to improving patient flexibility and privacy, the dashboard enables health workers to triage patients and focus their limited resources on the highest risk patients. Preliminary studies suggest that DATs can improve adherence in multiple disease settings \cite{haberer2017real, corden2016medlink, sabin2015improving}, prompting its use  and evaluation for managing TB adherence \cite{garfein2015feasibility,liu2015effectiveness}. The WHO has even published a guide for the proper implementation of the technology in TB care \cite{world2017handbook}.

\textit{In this paper, we study how the wealth of longitudinal data produced by DATs can be used to help health workers better triage TB patients and deliver interventions to boost overall adherence of their patient cohort.} The data we analyze is from Mumbai, India and comes from a partnership with the City TB Office of Mumbai; they have implemented a DAT by which patients prove adherence through daily toll-free calls. The DAT system was implemented with technical support from the healthcare technology company Everwell \cite{Everwell} and is known as 99DOTS \cite{cross201999dots}. In fact, Everwell supports implementations of 99DOTS throughout India where there were an estimated 2.7 million cases of TB in 2017 \cite{GTR18}. In Mumbai, patients enrolled in 99DOTS currently receive interventions according to the following general guidelines. If they have not taken their medication by the afternoon, they (and their health worker) receive a text message reminder. If the patient still does not take their medication by some time later, the worker will call the patient directly. Finally, if a patient simply does not respond to these previous interventions after some number of days, they may be personally visited by a health worker. Note that many of these patients live in low-resource communities where each health worker manages tens to hundreds of patients; far more than they can possibly visit in a day. Thus, models that can identify patients at risk of missing doses and prioritize interventions by health workers are of paramount importance.

At first glance, the problem of predicting whom to target for an intervention appears to be a simple supervised machine learning problem. Given data about a patient's medication adherence through their calls to the 99DOTS system, one can train a machine learning model to predict whether they will miss medication doses in the future.  However, such a model ignores the concurrent interventions from health workers as the data was collected, and can lead to incorrect prioritization decisions even when it is highly accurate. For instance, we might observe that missed doses are followed by a period of medication adherence: this does not mean that people with missed doses are more likely to take medication, but most likely that there was an intervention by a health worker after which the patient restarted their medication. 

Thus, for \emph{prescribing} interventions, we need to disentangle the effect of manual interventions from other underlying factors that result in missing a dose. However, since this data was collected via an extensive rollout to real patients, the data contains the effects of interventions carried out by health workers. As an additional challenge, health workers rarely record their interventions on the 99DOTS system, making it difficult to estimate their effects. While there is a well-developed literature on estimating heterogeneous treatment effects, standard techniques uniformly require knowledge of which patients received an intervention \cite{morgan2014counterfactuals,dehejia2002propensity,athey2016recursive,sutton2009convergent}. We note that such gaps will be common as countries eagerly adapt DAT systems in the hopes of benefiting low-income regions; to support the delivery of improved care, we must be able to draw lessons from this messy but plentiful data. 

Therefore, we introduce \textit{a general approach for learning from adherence data with unobserved interventions}, based on domain knowledge of the intervention heuristics applied by health workers. We construct a proxy for interventions present in the historical 99DOTS data and develop a model to help prioritize targets of interventions for health workers in different clinical scenarios:

\textbf{Modeling Daily Non-Adherence Risk.} We propose the prediction task: given adherence data up to a certain time period for patients not currently considered for intervention, predict risk of non-adherence in the next week. We then introduce machine learning models for this task, which enable health workers to accurately identify 21\% more high-risk patients and catch nearly 76\% more missed doses compared to the heuristics currently used in practice.

\textbf{Predicting success of treatment.} Next, we apply our framework to predict the final outcome at the end of the six-month treatment for a patient based on their initial adherence data. Like the previous  model, this can be useful for health workers to prioritize patients who are at risk of an unfavorable outcome, even though their adherence might be high. Additionally, since this prediction applies over the course of several months (rather than just one week in the previous task), this model can be useful for public health officials to better plan for TB treatment in their area, e.g. by assigning or hiring additional health workers. We show that our model can be used to achieve city-wide treatment outcome goals at nearly 40\% less cost than via baselines.

\textbf{Detecting Low-Call Favorable Outcome Patients.}
One critical challenge of the 99DOTS system is that almost 15\% of patients regularly take their doses as prescribed, but choose not to call. This causes the dashboard to falsely label those patients as HIGH risk, putting extra strain on health workers to validate those patients' adherence outside of 99DOTS. We uncover helpful insights including a simple rule that could allow health workers to identify such patients after only 7 days of adherence data has been collected, allowing workers to transition those patients onto different adherence monitoring technology.

\textbf{Decision Focused learning.} Finally, building on recent work in end-to-end decision-focused learning \cite{wilder2018melding}, we build a machine learning model which is tailored for a specific intervention planning problem. In the planning problem, workers must balance travel costs while predicting which patients will benefit most from interventions.  This example demonstrates how the modeling flexibility enabled by our approach allows us to fine-tune and extract additional gains for particular decision support tasks (in this case, a 15\% improvement over our earlier model).

With our proposed models, 99DOTS can leverage years of adherence data to better inform patient care and prioritize limited intervention resources. Additionally, the challenges we address are not unique to 99DOTS or TB adherence. DATs have been implemented for disease treatment regimens such as HIV and diabetes across the globe, and in each case health workers face the same challenge of prioritizing patient interventions. By enabling health workers to intervene before more missed doses, our model will directly contribute to saving the lives of those afflicted with TB and other diseases. That is why, though our model is not yet deployed, we are excited about our continued collaboration with the City TB Office of Mumbai and prospectively testing our model in the field.


\section{Related Work}
Outcomes and adherence research are well studied in the medical literature for a variety of diseases \cite{kardas2013determinants}. Traditionally, studies have attempted to identify demographic or behavioral factors correlated with non-adherence so that health workers can focus interventions on patients who are likely to fail. Tuberculosis in particular, given its lethality and prevalence in developing countries, has been studied throughout the world including in Ethiopia \cite{shargie2007determinants}, Estonia \cite{kliiman2010predictors}, and India \cite{roy2015risk}. Typically these studies gather demographic and medical statistics on a cohort, observe their adherence and outcomes throughout the trial, then retrospectively apply survival \cite{shargie2007determinants, kliiman2010predictors} or logistic regression \cite{roy2015risk} analysis to determine covariates predictive of failure. Newer work has improved classification accuracy via machine learning techniques such as Decision Trees, Neural Networks, Support Vector Machines and more \cite{kalhori2013evaluation, hussain2018predicting, sauer2018feature, mburu2018use}. However, the conclusions connecting predictors to risk are largely the same as in previous medical literature. While such studies have improved  patient screening at the time of diagnosis, they offer little knowledge about how risk changes \emph{during} treatment. In this work, we show how a patient's real-time adherence data can be used to track and predict risk changes throughout the course of their treatment. Previous studies likely did not address this question because accurately measuring patient adherence has historically been difficult. 

However, in recent years, new technologies have made measuring daily adherence feasible in the context of many diseases such as HIV or stroke. One such common device is an electronic pill bottle cap that records the date/time when the cap is removed. While some previous work has used electronic cap data to determine predictors of non-adherence \cite{platt2010can,pellowski2016alcohol,cook2017prospective}, almost no research has used the daily measurements made possible by the electronic cap to study changes in adherence over time. One study used the smart pillbox data to \emph{retrospectively} categorize patient adherence \cite{kim2018algorithm}, but our focus is on \emph{prospective} identification of patients at risk of missing doses before failures occur. As such devices enter mainstream use, machine learning techniques like the ones we propose will play an important role in the treatment of a wide spectrum of diseases.

Methodologically, our work is related to the large body of research that deals with estimating the causal impact of interventions from observational data \cite{morgan2014counterfactuals,dehejia2002propensity,athey2016recursive,sutton2009convergent}. Given appropriate assumptions, such techniques allow for valid inferences about counterfactual outcomes under a different policy for determining interventions. However, they crucially require exact knowledge of when interventions were carried out. This information is entirely absent in our setting, requiring us to develop new methods for handling \emph{unobserved} interventions in the training data.

\section{Data Description}
\label{DataDescription}
99DOTS provides each patient with a cover for every sleeve of pills that associates a hidden unpredictable phone number with each daily dose (note that one dose may consist of 2-5 pills). As patients expose pills associated with each dose, they expose one phone number per day. Each patient is instructed to place a toll-free call to the indicated number each day. 99DOTS counts a dose only if the patient calls the correct number for a given day. Due to the sensitivity of the health domain, all data provided by our partners was fully anonymized before we received it. The dataset contains over 2.1 million dose records for about 17,000 patients, served by 252 health centers across Mumbai from Feb 2017 to Sept 2018. \textbf{Table \ref{DataDescriptionTable}} summarizes the data. 
We now describe the available information in more detail.
\begin{table}[tbh]
\caption{Data Summary. *Doses per patient was calculated only on patients enrolled at least 6 months before Sept 2018.}
\begin{center}
\begin{tabular}{lr}
\toprule
 Metric & Count \\
\midrule
    Total doses recorded &      2,169,976 \\
    ---By patient call &        1,459,908 \\
    ---Manual (entered by health worker) &              710,068 \\
    Registered phones &       38,000 \\
    Patients &                16,975 \\
    Health centers &             252 \\
    Doses recorded per patient* & \\
    ---Quartiles &        57/149/188 \\
    ---Min/Mean/Max &       1/136/1409 \\
    Active patients per center per month &  \\
    ---Quartiles &        7/18/35 \\
    ---Min/Mean/Max &       1/25/226 \\
\bottomrule
\label{DataDescriptionTable}
\end{tabular}
\end{center}
\end{table}

\textbf{Patient Details. }
This is the primary record for patients who have enrolled with 99DOTS. The table includes demographic features such as weight-band, age-band, gender and treatment center ID. Also included are treatment start and end dates, whether treatment is completed or ongoing, and an "adherence string" which summarizes a patient's daily adherence. For patients who completed treatment, a treatment outcome is also assigned according to the standard WHO definitions \cite[p.~5]{WHODefandRepforTB13}. We label "Cured" and "Treatment Complete" to be favorable outcomes and "Died", "Treatment failed", and "Lost to follow-up" to be unfavorable outcomes. 

\textbf{Mapping phone numbers to patients. }
Patients must call from a registered phone number for a dose to be counted by the 99DOTS system. Patients can register multiple phones, each of which will be noted in the Phone Map table. We filtered out phones that were registered to multiple patients since they could not be uniquely mapped to patients. Also, patients who had \textit{any} calls from shared phones were filtered out to avoid analyzing incomplete call records. This removed <1\% of the patients from the data set.

\textbf{Call Log. }
The Call Log records every call received by 99DOTS, including from patients outside of Mumbai. It also includes "manual doses" marked by health workers. Manual doses allow workers to retroactively update a patient's adherence on the dashboard. For instance, if a patient missed a week of calls due to a cellular outage, the worker could update the record to account for those missed doses. We filtered the Call Log to only contain entries with patients and phones registered in Mumbai, then attached a Patient ID to each call by joining the filtered Call Log and Phone Map.

\textbf{Patient Log. }
Each time a health worker interacts with a patient's data in the 99DOTS dashboard, an automatic note is generated describing the interaction. The Patient Log records each such event, noting the type of action, Patient ID, health worker ID, the health worker's medical unit, what action was taken, and a timestamp. We did not calculate features from this table as they tended to be sparse. However, this table was used for calculating our training labels as described in \textbf{Section \ref{handlingInterventions}}.


\section{Unobserved Interventions}
\label{handlingInterventions}


The TB treatment system operates under tight resource limitations, e.g. one health worker may be responsible for more than 100 patients. Thus, it is critical that workers be able to accurately rank patient risk and prioritize interventions accordingly. 
Machine learning can be used to accomplish such risk ranking with promising accuracy, but it requires taking special care to understand how intervention resources were allocated in the existing data. 

Therefore, a key challenge is that users of the 99DOTS platform generally do not record interventions: workers may make texts, calls, or personal visits to patients to try to improve adherence, but these interventions are not routinely logged in the data. While far from ideal, such gaps are inevitable as countries with differing standards of reporting adopt DATs for TB treatment. Given the abundance of data created by DATs and their potential to impact human lives, we emphasize the importance of learning lessons in this challenging setting where unobserved interventions occur. We next resolve this challenge by formulating a screening procedure which identifies patients who were likely candidates for particular interventions. However, we first illustrate the pitfalls of training and using a risk model in this domain without our screening procedure.


Consider a naive model trained on the data as-is. Some of the data will be influenced by the historical interventions carried out by health workers. Thus, such a model will learn how to predict patient adherence \emph{given existing worker behaviors}. 
Now consider the model's intended use, namely to recommend a new prioritization of limited resources based on risk. Then in deployment, some patients who would have received interventions under the historical policy would be judged not to require intervention by the new model. While such prioritization is desirable under resource constraints, naive models which ignore the impact of interventions in the dataset can actually \emph{worsen} patient outcomes. For instance, assume we use the naive model to make a prediction about the patient from \textbf{Section \ref{Introduction}} who had a week of missed doses, an intervention, then a week of good adherence. By correctly predicting this patient's good adherence the naive model would recommend no intervention -- but this patient's good adherence is \textit{contingent on} the hidden intervention in the data. \emph{Hence, the naive model will take resources away from exactly the patients who would benefit most.} To avoid such pitfalls arising from unobserved interventions, we must train and evaluate on data that is not influenced by such intervention effects. We now describe our general method for reshaping data around intervention effects to build valid models.  


\textbf{Intervention Proxy. }
Our goal is to use the available data to formulate a proxy for when an intervention is likely to have occurred, so that we can train our models on data points which are unaffected by interventions. \textit{The key is to identify a conservative estimate for where interventions occur to ensure that data with intervention signals are not included.} First, we draw a distinction between different types of health worker interventions. Specifically, we consider a house visit to be a "resource-limited" intervention since workers cannot visit all of their patients in a timely manner. Generally, this is a last resort for health workers when patients will not respond to other methods. Alternatively, we consider calls and texts to be "non-resource-limited" interventions since they could feasibly be made on a large number of patients at very low cost. Note that the naive model in the previous section \textit{could} make valid recommendations for actions \textit{in addition} to normal health worker behaviors. For this reason, we develop a proxy only for resource-limited interventions since  non-resource-limited interventions come virtually for "free".

To formulate our proxy, we first searched for health worker guidelines for carrying out house visits. The 2005 guide by India's Revised National Tuberculosis Control Program (RNTCP) \cite{rntcpTOG2005} required that workers deliver a house visit after a single missed dose, but updated guides are far more vague on the subject. Both the most recent guide by the WHO \cite{world2017handbook} and by the RNTCP \cite{rntcpTOG2016} leave house visits up to the discretion of the health worker. However, through discussions in Mumbai we learned that health workers prioritize non-adherent patients for resource-limited interventions such as house visits. Thus, we formulated our proxy based on the adherence dashboard seen by health workers.

The 99DOTS dashboard gives a daily "Attention Required" value for each patient. First, if a patient has an entry in the Patient Log (i.e. provider made a note about the patient) in the last 7 days they are automatically changed to "MEDIUM" attention, but this rule affects <1\% of the labels. The remaining 99\% of labels are as follows: If a patient misses 0 or 1 doses in the last 7 days, they are changed to "MEDIUM" attention, whereas if they miss 4 or more they are changed to "HIGH" attention. Patients with 2-3 missed doses retain their attention level from the previous day. As our conservative proxy, we assumed that only "HIGH" attention patients were candidates for resource-limited interventions since the attention level is a health worker's primary summary of recent patient adherence. This "Attention Required" system for screening resource-limited interventions is generalizable to any daily adherence setting; one need only to identify the threshold for a change to HIGH attention. 

With this screening system, we can identify sequences of days during which a patient was a candidate for a resource-limited intervention, and subsequently avoid using signal from those days in our training task. We accomplish this with our formulation of the real-time risk prediction task as follows.

Consider a given set of patients on the dashboard of a health worker at day $t$. Each patient will have an "Attention Required" value in \{MEDIUM, HIGH\} representing their risk for that day. Over the course of the next week up to $t+7$, we will observe call behavior for each patient and so the attention for each patient may also change each day. Between $t+1$ and $t+7$, any patient that is at HIGH on a given day may receive a resource-limited intervention while those at MEDIUM may not. Note that a change from MEDIUM to HIGH on day $t_i$ where $t+1 \leq t_{i} \leq t+7$  means that a patient missed 4 doses over days $[t_{i}-6, t_{i}]$. Patients at HIGH attention are already known to the health worker, so the goal for our ML system is to help prevent transitions from MEDIUM to HIGH by predicting which patients are at greatest risk before the transition occurs and allowing a health worker to intervene early.


We formalize our prediction task as follows. For each patient who is MEDIUM at time $t$, use data from days $[t-6, t]$ to predict whether or not they change to HIGH at any time $t_{i}$ where $t+1 \leq t_{i} \leq t+7$. We now demonstrate that, with our intervention proxy, resource-limited intervention effects cannot effect labels in this formulation. First, if a patient stays at MEDIUM for all $t_{i}$, then the label is 0. Since the patient was at MEDIUM for all $t_{i}$, our proxy states that no resource-limited intervention took place between our prediction time $t$ and the time that produced the label, $t+7$. Second, if a patient changes from MEDIUM to HIGH on day $t_{i}$, then on day $t_{i}$ we establish that the label is 1. By our proxy, any resource-limited intervention effect must happen in $[t_{i} + 1, t + 7]$, since attention is established at the end of a day $t_{i}$. So again, we have that no resource-limited intervention took place between our prediction time $t$ and the time that produced the label, $t_{i}$. 

Since we ensure that no resource-limited interventions happen between our prediction time and the time the label is generated, we ensure that intervention effects cannot influence our labels. Now, if we predict that a patient will have good adherence we can safely recommend no intervention since our combined screening and training method guarantees that their good adherence \textit{is not contingent on} an intervention. Thus our classifier is suited to make predictions that prioritize resource-limited interventions.

\textit{Despite messy data affected by unobserved interventions, this conservative, general proxy generates clean data without interventions.} In the next section, we show how this approach leads to significant improvements in prediction performance and creates valid recommendations to enable interventions among patients at immediate risk of becoming non-adherent.

\section{Real-Time Risk Prediction}
We now build a model for the prediction task formalized in \textbf{Section \ref{handlingInterventions}} which leverages our intervention screening proxy. Our goal was to develop a model corresponding to the health worker's daily task of using their patients' recent call history to evaluate adherence risk with the goal of scheduling different types of interventions. Better predictions allow workers to proactively intervene with more patients before they miss critical doses.

\textbf{Sample Generation. }
We started with the full population of 16,975 patients and generated training samples from each patient as follows. We considered all consecutive sequences of 14 days of call data where the first 7 days of each sequence were non-overlapping. We excluded each patient's first 7 days and the last day of treatment to avoid bias resulting from contact with health workers when starting or finishing treatment. We then took two filtering steps. First, we removed samples where the patient had more than 2 doses manually marked by a provider during the input sequence since these patients likely had contact with their provider outside of the 99DOTS system. Second, we removed samples in which the patient did not miss any doses in the input sequence. These samples made up the majority of data but included almost no positive (HIGH risk) labels, which distorted training. Further, positive predictions on patients who missed 0 doses are unlikely to be useful; no resource-limited intervention can be deployed so widely that patients with perfect recent adherence are targeted. The above procedure generated 16,015 samples (2,437 positive). 

\textbf{Features. }
Each sample contained a time-series of call data and static features. The time series included two sequences of length 7 for each sample. First was a binary sequence of call data (1 for a call or manual dose and 0 for a miss.) The second sequence was a cumulative total of all doses missed up to that day, considering the patient's full history in the program. The static features included four demographic features from the Patient Table: weight-band, age-band, gender, and treatment center ID. Additional features were engineered from the patient Call Logs and captured a patient's \textit{behavior} rather than just their adherence. For example, does the patient call at the same time every morning or sporadically each day? This was captured through the mean and variance of the call minute and hour. Other features included number of calls, number of manual doses, mean/max/variance of calls per day as well as days per call. We also included analogous features which used only \textit{unique} calls per day (i.e. calls \textit{to} unique phone numbers), or ignored manual doses. This process resulted in 29 descriptive features.

\begin{figure}
  \includegraphics[width=0.5\textwidth]{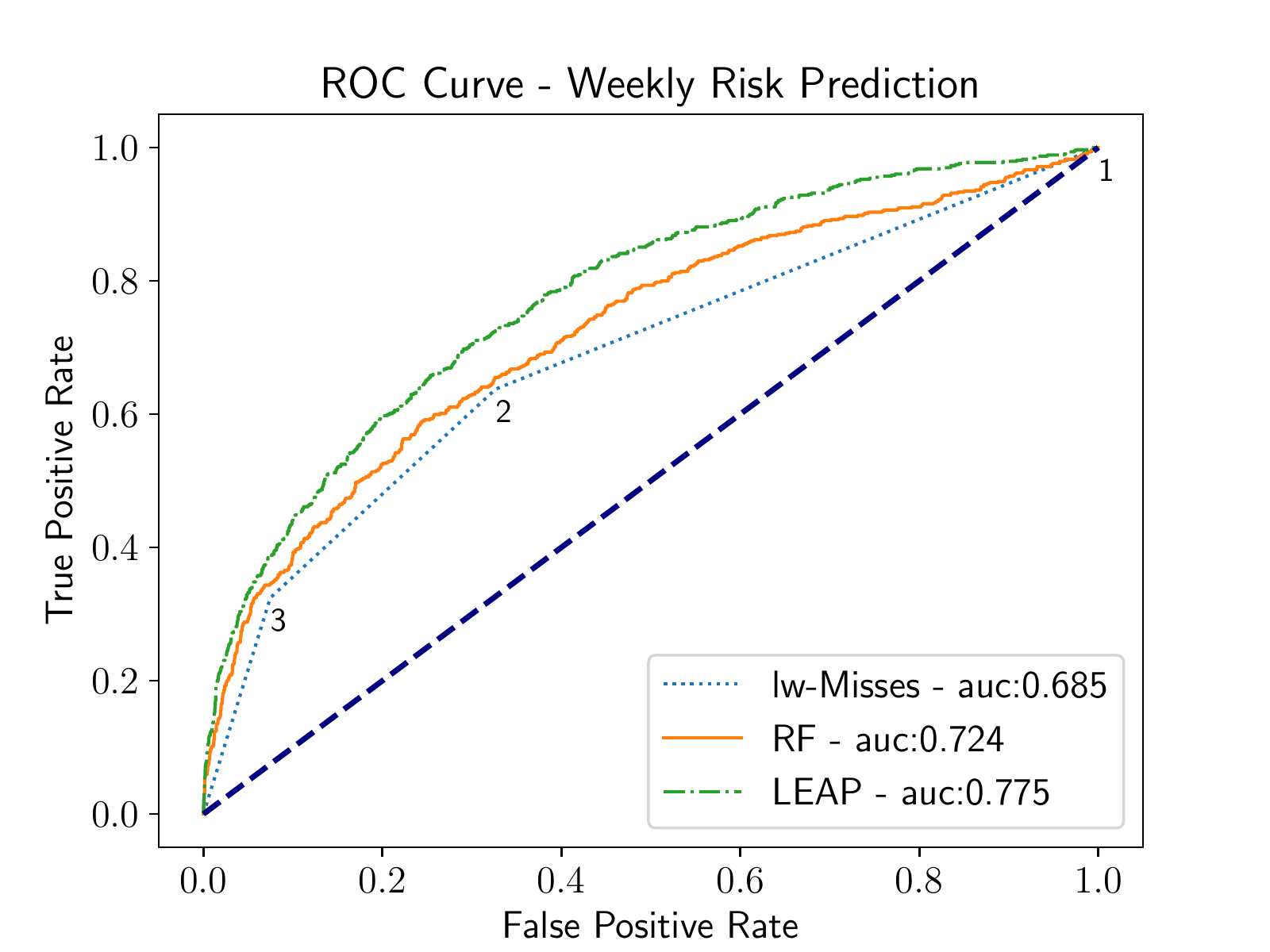}
  \caption{ROC Curve for the weekly risk prediction task comparing the missed call baseline (blue), Random Forest (yellow) and LEAP (green). Numbers under the blue curve give thresholds used to calculate the baseline's ROC curve.}
  \label{WeeklyROC}
\end{figure}

\textbf{Models. } We first tested standard models which use only the static features: linear regression, a random forest \cite{scikit-learn} (with 100 trees and a max depth of 5), and a support vector machine. The random forest performed best, so we exclude the others for clarity. To leverage the time series data we also built a deep network, named LEAP (Lstm rEal-time Adherence Predictor), implemented with Keras \cite{chollet2015keras} which takes both the time series and static features as input. LEAP has two input layers: 1) a LSTM with 64 hidden units for the time series input and 2) a dense layer with 100 units for the static feature input. We concatenated the outputs of these two layers to feed forward into another dense layer with 16 units, followed by a single sigmoid activation unit. We used a batch size of 128 and trained for 20 epochs.

\textbf{Model Evaluation. }
To evaluate models we randomized all data then separated 25\% as the test set. We used 4-fold grid search to determine the best model parameters. To deal with class imbalance, we used SMOTE to over-sample the training set \cite{chawla2002smote} implemented with the Python library imblearn \cite{JMLR:v18:16-365}. We also normalized features as percentiles using SKLearn \cite{scikit-learn} which we found empirically to work well. The baseline we compared against was the method used by the existing 99DOTS platform to asses risk, namely doses missed by the patient in the last week (lw-Misses).

\textbf{Figure \ref{WeeklyROC}} shows the ROC curve of our models vs. the baseline. The random forest narrowly outperforms the baseline and LEAP clearly outperforms both. However, to evaluate the usefulness of our methods over the baseline, we consider how each method might be used to plan house-visit interventions. Since this is a very limited resource, we set the strictest baseline threshold to consider patients for this intervention; that is 3 missed calls. Fixing the FPR of this baseline method, \textbf{Table \ref{aucWeeklythresh3Table}} shows how many more patients in the test set would be reached each week by our method (as a result of its higher TPR) as well as the improvement in number of missed doses caught. To calculate missed doses caught, we count only missed doses that occur before the patient moves to HIGH risk. \emph{Our model catches 21.6\% more patients and 76.5\% more missed doses, demonstrating substantially more precise targeting than the baseline.}

\begin{table}
\begin{center}
\caption{LEAP vs. Baseline - Missed Doses Caught}
\begin{tabular}{lrr}
\toprule
      Method &  True Positives &  Doses Caught \\
\midrule
    Baseline &      204 &    204 \\
    LEAP &      248 &    360 \\
 Improvement &        \textbf{21.6\%} &      \textbf{76.5\%} \\
\bottomrule
\label{aucWeeklythresh3Table}
\end{tabular}
\caption*{LEAP vs.\ baseline for catching missed doses with a fixed false positive rate. Our method learns behaviors indicative of non-adherence far earlier than the baseline, allowing for more missed doses to be prevented.}
\end{center}
\end{table}
\begin{table}
\begin{center}
\caption{LEAP vs. Baseline: Additional Interventions}
\begin{tabular}{rrrr}
\toprule
  TPR &  Baseline FPR &  LEAP FPR &  Improvement \\
\midrule
 75\% &      50\% &      35\% &    \textbf{30}\% \\
 80\% &      63\% &      41\% &    \textbf{35}\% \\
 90\% &      82\% &      61\% &    \textbf{26}\% \\
\bottomrule
\label{aucWeeklythresh1Table}
\end{tabular}
\caption*{LEAP vs.\ baseline for implementing new interventions. At any TPR LEAP improves over the baseline FPR, allowing for more precisely targeted interventions.}
\end{center}
\end{table}

\textbf{Table \ref{aucWeeklythresh1Table}} shows that our model also outperforms the baseline as both the true positive rate (TPR) and FPR increase, showcasing our model's greater discriminatory power. This is useful for non-resource-limited interventions such as calls or texts. Recall, that our screening procedure does not apply to this type of intervention, so our predictions may only recommend \textit{additional} interventions. It is important that additional interventions be carefully targeted since repeated contact with a given patient reduces the efficacy of each over time \cite{demonceau2013identification}. This highlights the value of the greater precision offered by our model, since simply blanketing the entire population with calls and texts is likely counterproductive.

\textbf{Interpretability. }
Our model has the potential to catch more missed doses than current methods. However, these gains cannot become reality without health workers on the ground delivering interventions based on the predictions. Interpretability is thus a key factor in our model's usefulness because health workers need to understand \textit{why} our model makes its predictions to trust the model and integrate its reasoning with their own professional knowledge.

\begin{figure}[tp]

\subfloat[SHAP values for LEAP's dense layer features for a high-risk sample ($\geq$0.5).]{%
  \includegraphics[width=0.5\textwidth]{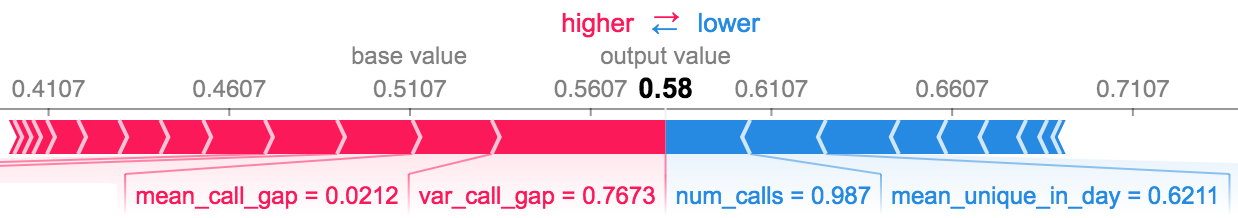}%
}

\subfloat[SHAP values for LEAP's LSTM layer input for 4 samples.]{%
  \includegraphics[width=0.5\textwidth]{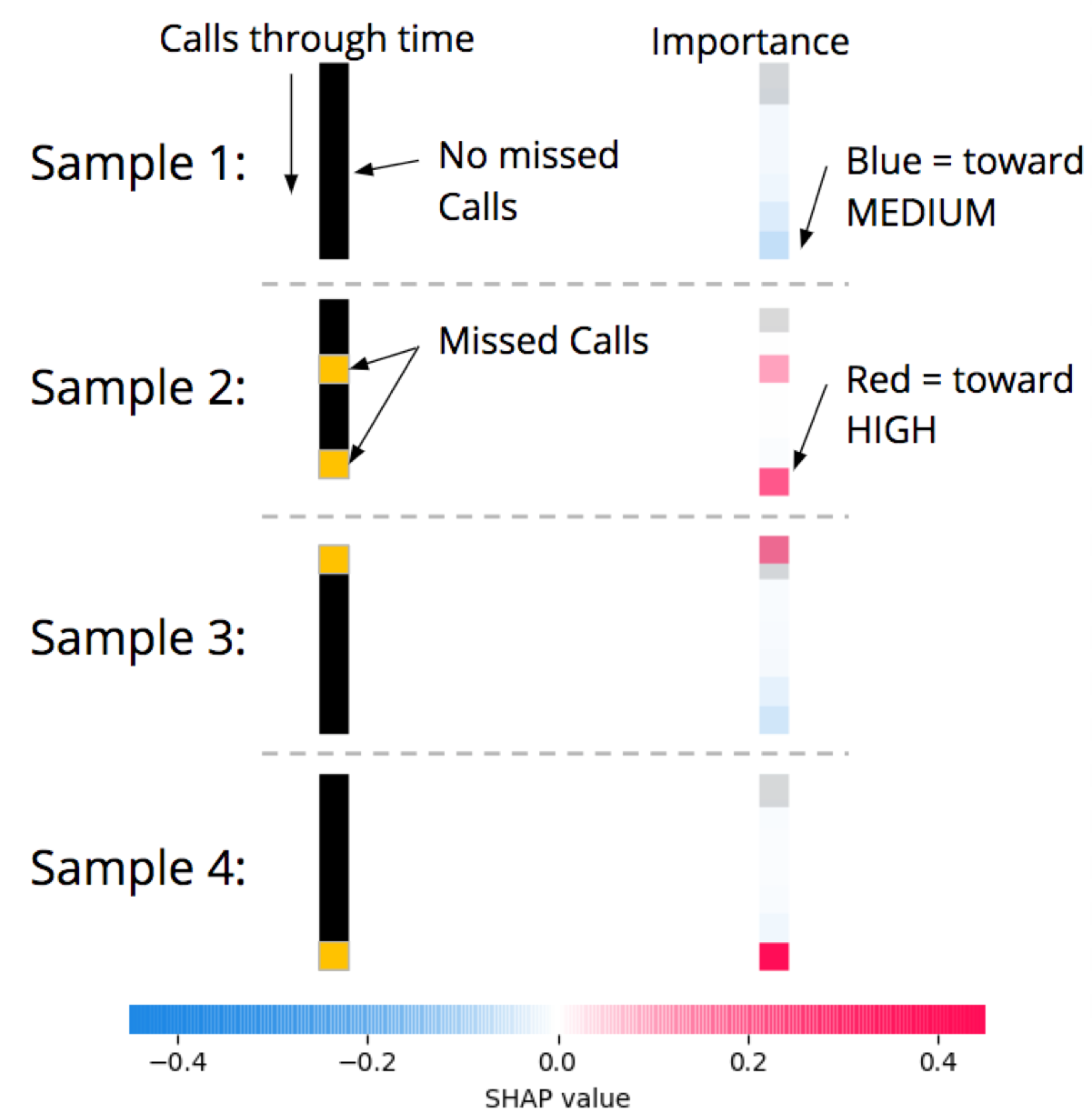}%
}

\caption{Visualization of the (a) dense layer and (b) LSTM layer of our weekly risk prediction model. Red values correspond to inputs that push predictions toward output of 1; blue values push toward output of 0.}
\label{ShapFig}
\end{figure}

However, the best predictive performance was achieved with LEAP, a black-box network, rather than a natively interpretable model like linear regression. Accordingly, we show how a visualization tool can help users draw insights about our model's reasoning. We used the SHapley Additive exPlanations (SHAP) python library, which generates visualizations for explaining machine learning models \cite{shap}. \textbf{Figure \ref{ShapFig}a} shows how static features influence our model's prediction, where red features push predictions toward 1 (HIGH) and blue toward 0 (MEDIUM). Recall that features are scaled as percentiles. In the blue, we see that this patient makes an above-average number of calls each week pushing the prediction toward 0. However, in the red we see that this patient has a \textit{very low average} but a \textit{high variability} in time between calls. These features capture that this patient missed two days of calls, then made three calls on one day in an attempt to "back log" their previous missed calls. Our model learned that this is a high-risk behavior.



\textbf{Figure \ref{ShapFig}b} shows four different samples as input to the LSTM layer of our model. The left shows the binary input sequence as colored pixels where black is a call and yellow is a missed call. On the right are SHAP values corresponding to each day of adherence data, and grey denotes the start of the call sequence. We see that the model learned that calls later in the week carry more weight than calls earlier in the week. In Sample 1, the bottom two pixels (the most recent calls) have blue SHAP values while the other pixels have SHAP value close to 0. In Sample 3, a single missed call at the beginning of the week combined with a call made at the end of the week result in essentially cancelling SHAP values. Sample 4 also has one missed call but on the last day of the week, resulting in a net positive SHAP value.

This visualization technique provides intuitive insights about the rules learned by our model. In deployment, workers could generate these visualizations for any sample on the fly in order to aid their decision-making process.

\section{Outcome Prediction}
\label{Outcome}
Next we investigate how adherence data can be used to predict final treatment outcome. Traditional TB treatment studies model outcomes only as they relate to patient covariates such as demographic features. Exploiting daily real-time adherence data provided by DATs, we investigate how using the first $k$ days of a patient's adherence enables more accurate, personalized outcome predictions. Note that intervention effects are still present in this formulation. However, our screening procedure will not apply since we predict over a period of several months, during which virtually all patients would have had repeated in-person contact with health workers. 

\textbf{Sample Generation and Features. }
\label{OutcomeDataGeneration}
We formalize the prediction task as follows: given the first $k$ days of adherence data, predict the final binary treatment outcome. We considered "Cured" and "Treatment Complete" as favorable outcomes and "Died", "Lost to follow-up", and "Treatment Failure" as unfavorable outcomes. We only include patients who were assigned an outcome from these categories. Further, since patients with the outcome "Died" or "Lost to follow-up" exit the program before the full 6 months of treatment, we removed those who were present for less than $k+1$ days. Finally, we removed patients who had more than half their first k days marked as manual doses. This tended to improve prediction performance which we conjecture is related to our observation that practices for reporting manual doses varied by health center -- making the "meaning" of a manual dose ambiguous across samples with respect to outcome. Our final dataset contained 4167 samples with 433 unfavorable cases. 

Through discussions in Mumbai, we learned that health workers often build a sense of a patient's risk of an unfavorable outcome within their first month of treatment. To model this process, we set k=35 for our prediction task, capturing the first month of each patient's adherence after enrollment in 99DOTS. (Note that this is not a general rule for health workers, but simply served as a motivation for our choice of k in this task.) Both the static features and the sequence inputs were the same as calculated for the weekly prediction task, but now taken over the initial 35 days. We included two versions of the health worker baseline: missed doses in the last week (lw-Misses) and total missed doses in 35 days (t-Misses).

\textbf{Model Evaluation. }
\label{OutcomeModelEval}
We used the same models, grid search design, training process, and evaluation procedure as before. For the Random Forest we used 150 trees and no max depth. For LEAP, we used 64 hidden units for the LSTM input layer, 48 units for the dense layer input, and 4 units in the penultimate dense layer.

\begin{figure}
  \includegraphics[width=0.5\textwidth]{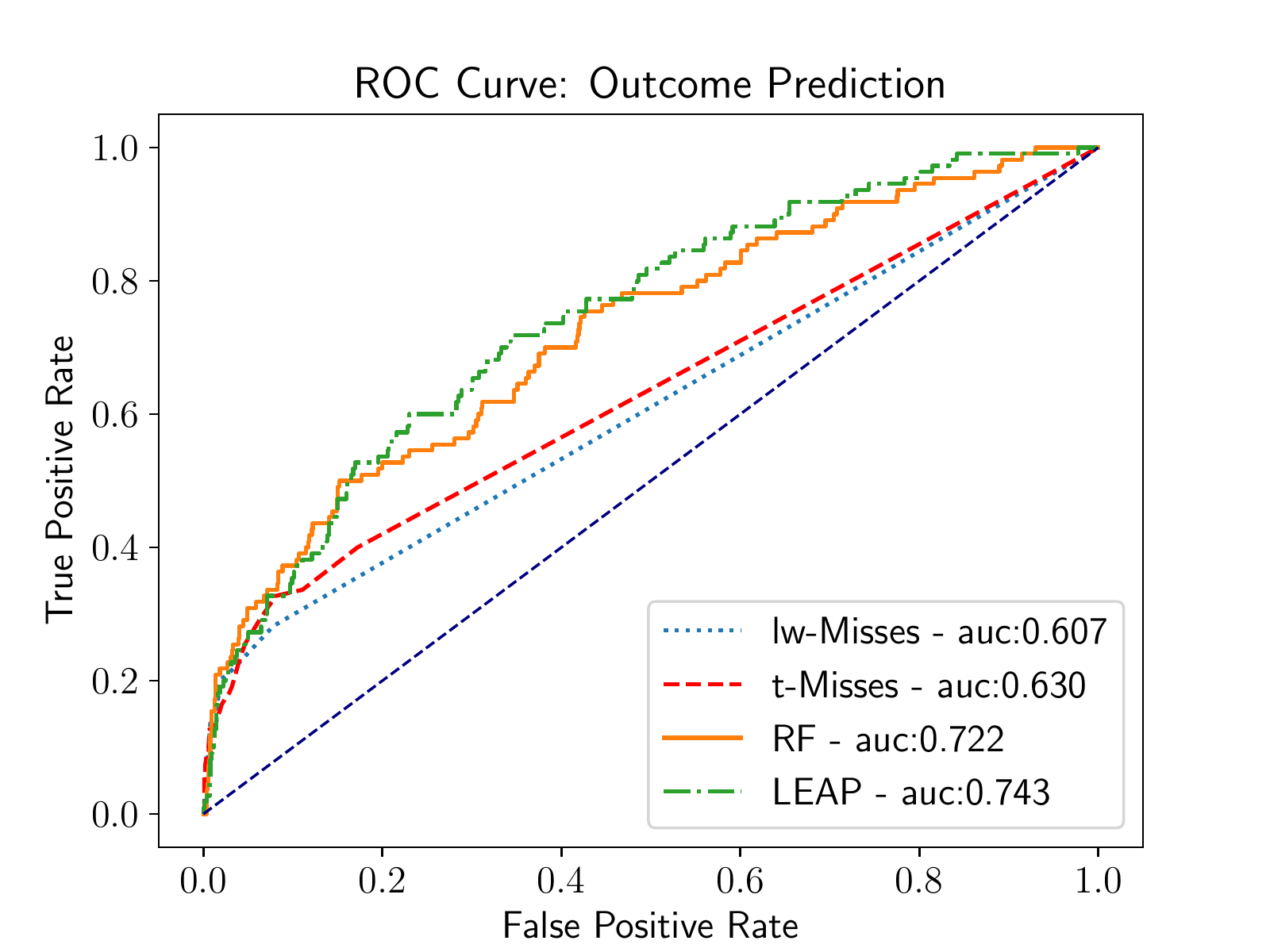}
  \caption{ROC curves for outcome prediction models.}
  \label{OutcomeROCunfiltered}
\end{figure}

\textbf{Figure \ref{OutcomeROCunfiltered}} shows ROC curves for each model. Even the very simple baseline of counting the calls made in the last 7 days before the 35 day cutoff is fairly predictive of outcome suggesting that the daily data made available by DATs is valuable in evaluating which patients will fail from TB treatment. Our ML models display even greater predictive power, with LEAP performing the best, followed closely by the random forest. We highlight how LEAP's predictive power could help officials minimize the costs necessary to reach medical outcome goals for their city. For example, say Mumbai launches a new initiative to catch 80\% of unfavorable outcomes (true positives in \textbf{Figure \ref{OutcomeROCunfiltered}}) by hiring new health staff. Over the ~17,000 patients in Mumbai, where 10\% have unsuccessful outcomes as in our test set, an 80\% catch rate requires saving 1360 patients. Using either baseline, achieving the 80\% TPR requires a FPR of 70\%, i.e., hiring additional staff to support \textit{10710} total patients in this example scenario. However, using LEAP only incurs a FPR of 42\%, translating to 6426 total patients. Recall that in Mumbai, the average health worker cares for about 25 patients. At a yearly starting salary of \rupee216,864 \cite{tbhvSalary} (or \$3026) our model would yield \rupee37M in saved costs (or ~\$525,000) per year.

\section{Detecting Low-Call Favorable Outcome Patients}
\label{LCFO}
One additional critical challenge of the 99DOTS system is that some patients regularly take their doses as prescribed, but choose not to call. So according to the dashboard, they \textit{seem} to be missing doses and would be classified as HIGH risk by 99DOTS and by LEAP alike, but in reality they should be MEDIUM risk. In fact, nearly 15\% of patients who had an outcome assigned as in section \ref{DataDescription} called on less than 25\% of days of their treatment, but had a favorable outcome. We refer to these patients as low-call favorable outcome (LCFO). We want to learn to identify these LCFO patients so that we do not falsely rank them as HIGH risk despite the fact that they are not calling. We also would like to be able to identify these patients early on in treatment so that they may be reassigned to an adherence monitoring method that better suits them. 

\textbf{Sample Generation and Features. }
We formulate this as a binary prediction task as follows: given the first $k$ days of adherence data, predict whether or not the patient will \textit{both} call on less than 25\% of days from day $k+1$ onward \textit{and} have a favorable outcome. We included only patients who were assigned an outcome as in Section \ref{DataDescription} and who had at least $k+7$ days of adherence data. To detect LCFO status as early as possible, we set $k=7$. Thus our final dataset contained 7265 patients, of which 1124 were positive. Note that this population was larger than that of our outcome prediction task because 1) patients were required to be in the program for less time and 2) patients were not removed for having too many manual doses since we found this to correlate with being LCFO.

Both the static features and the sequence inputs were the same as calculated for the outcome prediction task, but this time taken over the initial 7 days. We included the health worker baseline of missed doses in the last week (lw-Misses), a random forest trained only on demographic or "0-day" data (RF 0-day), a simple baseline which counts the number of manual doses in the last week (lw-Manual), a random forest trained on all non-sequence features over the initial 7 days (RF), and LEAP trained on all features and sequences.

\textbf{Model Evaluation. }
\label{LCFOModelEval}
We used the same models, grid search design, training process, and evaluation procedure as the previous two formulations. For RF 0-day we used 300 trees and a max depth of 10. For RF we used 200 trees and a max depth of 10. For LEAP, we used 200 hidden units for the LSTM input layer, 1000 units for the dense layer input, and 16 units in the penultimate dense layer.

\begin{figure}
  \includegraphics[width=0.5\textwidth]{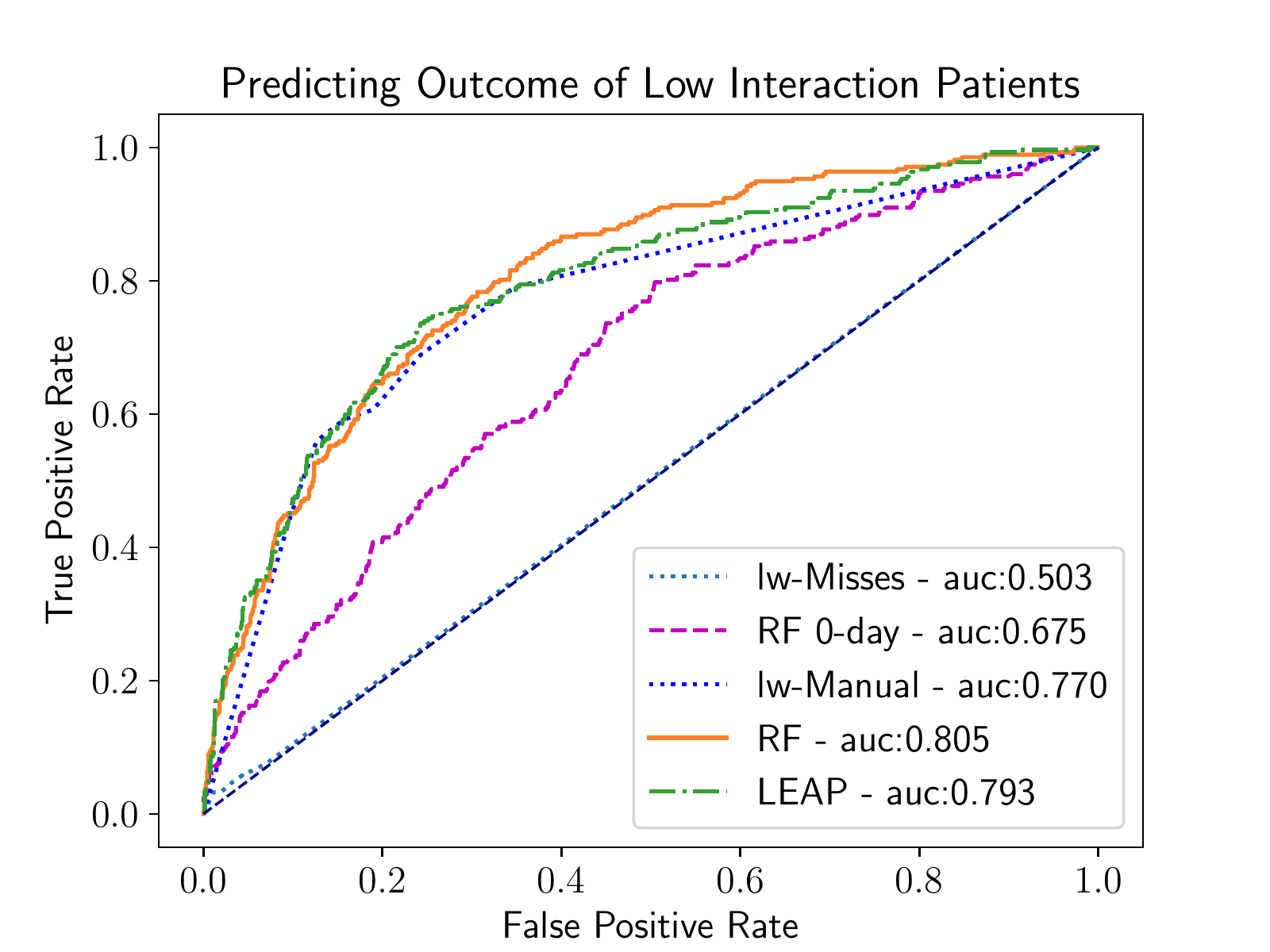}
  \caption{ROC curves for LCFO prediction models. }
  \label{OutcomeROC_LCFO}
\end{figure}

\textbf{Figure \ref{OutcomeROC_LCFO}} shows ROC curves for each model. Interestingly, for this task the lw-Misses baseline has almost no predictive power -- note that its ROC curve is essentially the line $y=x$. Conversely notable is the performance of the lw-Manual heuristic which simply counts the number of manual doses marked in the first 7 days for each patient. This simple heuristic has almost equivalent predictive power of our machine learning models. This is a valuable insight for health workers which suggests that if the worker is already manually marking doses for a patient early in their treatment, the patient is likely to continue to be disengaged with the system in the long term and should be considered for different adherence technology. The RF 0-day model has decent predictive power, though closer inspection reveals that most of this power is encoded in the treatment center ID -- that is, LCFO patients tend to be concentrated at certain treatment centers. This insight merits closer inspection by supervisors about why patients in certain regions tend to be disengaged with 99DOTS but still consuming pills. The RF and LEAP models both perform slightly better than the lw-Manual baseline but similarly to each other, suggesting that the adherence sequence structure does not encode additional information for this prediction task. These insights could improve processes by 1) helping to identify hotspot regions of LCFO patients, after which supervisors might investigate the underlying reason and adjust treatment accordingly at those centers and 2) the lw-Manual baseline, after \textit{only 7 days} of dosage data, could give health workers a simple rule for identifying LCFO patients that should switch to different adherence technology.

\section{Decision Focused Learning}
We now explore a case study of how our LEAP model can be specialized to provide decision support for a particular intervention. We exploit end-to-end differentiability of the model to replace our earlier loss function (binary cross-entropy) with a performance metric tailored to the objective and constraints of specific decision problem. To accomplish this end-to-end training, we leverage recent advances in \emph{decision-focused learning}, which embeds an optimization model in the loop of machine learning training \cite{wilder2018melding,donti2017task}. 

We focus on a specific optimization problem that models the allocation of health workers to intervene with patients who are at risk in the near future. This prospective intervention is enabled by our real-time risk predictions and serves as an example of how our system can enable proactive, targeted action by providers. However, we emphasize that our system can be easily modified to capture other intervention problems. Such flexibility is one benefit to our technical approach, which allows the ML model to \emph{automatically} adapt to the problem specified by a domain expert. 

Our optimization problem models a health worker who plans a series of interventions over the course of a week. The health worker is responsible for a population of patients across different locations, and may visit one location each day. We use location identifiers at the level of the TB Unit since this is the most granular identifier which is shared by the majority of patients in our dataset. Visiting a location allows the health worker to intervene with any of the patients at that location. The optimization problem is to select a set of locations to visit which maximizes the number of patients who receive an intervention \emph{on or before the first day they would have missed a dose}. We refer to this quantity as the number of \emph{successful interventions}, which we choose as our objective for two reasons. First, it measures the extent to which the health worker can proactively engage with patients before adherence suffers. Second, this objective only counts patients who start the week at MEDIUM attention and receive an intervention before they could have transitioned to HIGH, dovetailing with our earlier discussion on avoiding unobserved interventions in the data. This extends our earlier intervention proxy to handle day-by-day rewards.

We now show how this optimization problem can be formalized as a linear program. We have a set of locations $i = 1...L$ and patients $j = 1...N$ where patient $j$ has location $\ell_j$. Over days of the week $t = 1...7$, the objective coefficient $c_{jt}$ is 1 if an intervention on day $t$ with patient $j$ is successful and 0 otherwise. Our decision variable is $x_{it}$, and takes the value 1 if the health worker visit location $i$ on day $t$ and 0 otherwise. With this notation, the final LP is as follows:
\begin{align*}
    \max_{x} &\sum_{t= 1}^7 \sum_{i = 1}^L x_{it} \left(\sum_{j: \ell_j = i}c_{jt}\right)\\
    \text{s.t. }&\sum_{i = 1}^L x_{it} \leq 1, t = 1...7\\
    &\sum_{t = 1}^7 x_{it} \leq 1, i = 1...L\\
    & 0\leq x_{it} \leq 1 \quad \forall i,t
\end{align*}
where the second constraint prevents the objective from double-counting multiple visit to a location. We remark that the feasible region of the LP can be shown to be equivalent to a bipartite matching polytope, implying that the optimal solution is always integral. 

The machine learning task is to predict the values of the $c_{jt}$, which are unknown at the start of the week. We compare three models. First, we extend the lw-Misses baseline to this setting by thresholding the number of doses patient $j$ missed in the last week, setting $c_{jt} = 0$ for all $t$ if this value falls below the threshold $\tau$ and $c_{jt} = 1$ otherwise. We used $\tau = 1$ since it performed best. Second, we trained our LEAP system directly on the true $c_{jt}$ as a binary prediction task using cross-entropy loss. Third, we trained LEAP to predict $c_{jt}$ using performance on the above optimization problem as the loss function (training via the differentiable surrogate given by \cite{wilder2018melding}). We refer to this model as LEAP-Decision. 

We created instances of the decision problem by randomly partitioning patients into groups of 100, modeling a health worker under severe resource constraints (as they would benefit most from such a system). We included all patients, including those with no missed doses in the last week, since the overall resource allocation problem over locations must still account for them.

\begin{figure}
    \centering
    \includegraphics[width=1.5in]{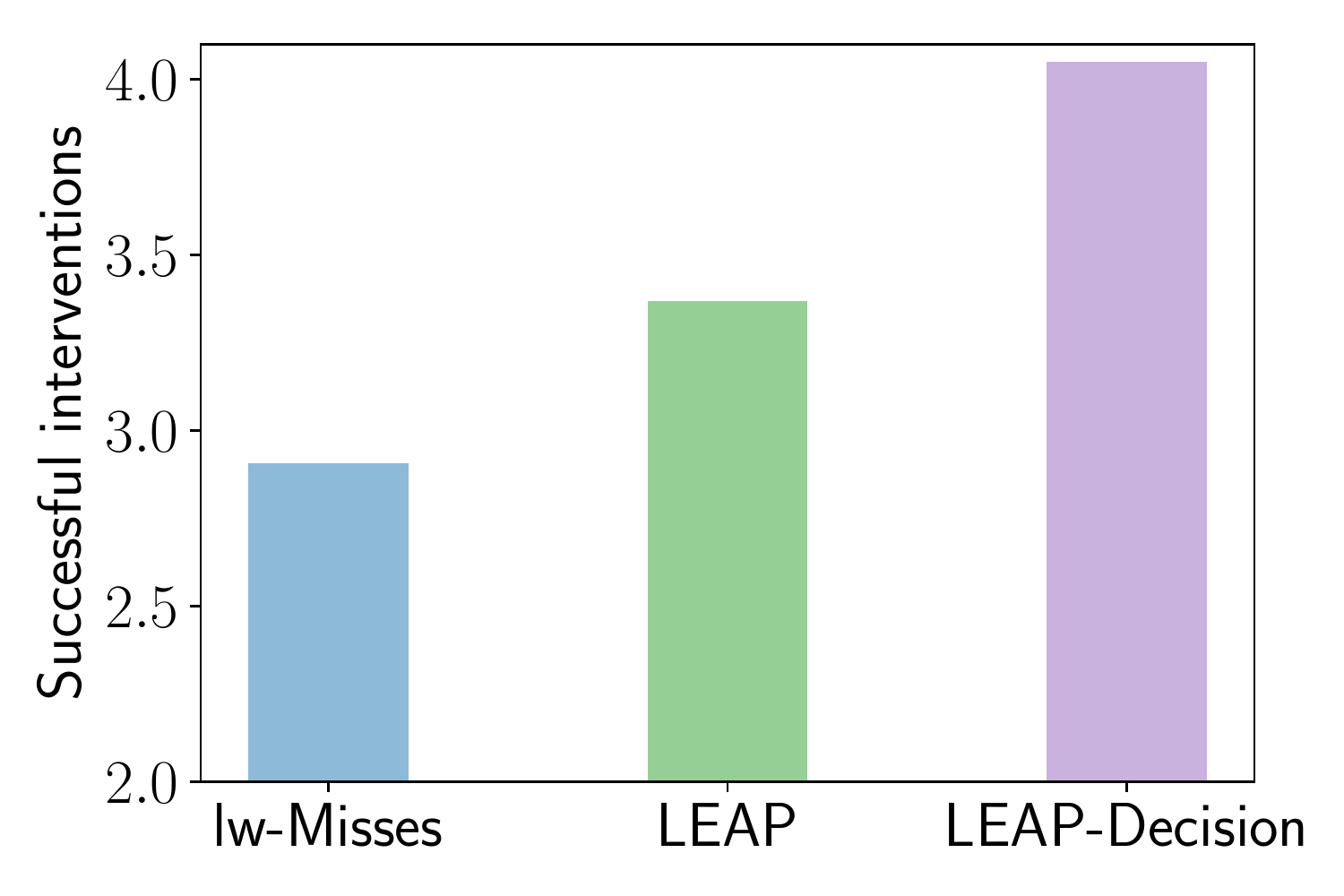}
    \includegraphics[width=1.5in]{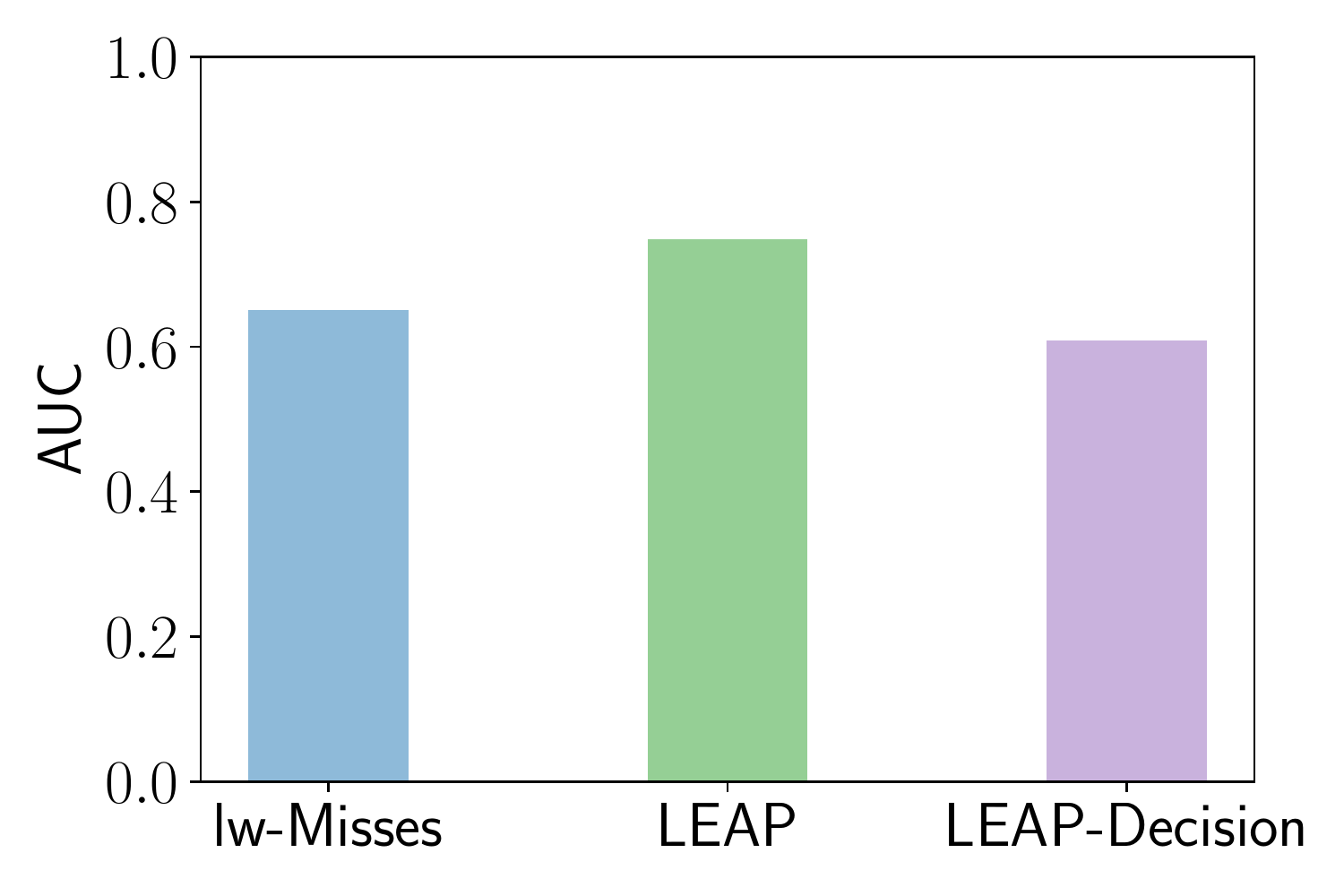}\\    \includegraphics[width=1.5in]{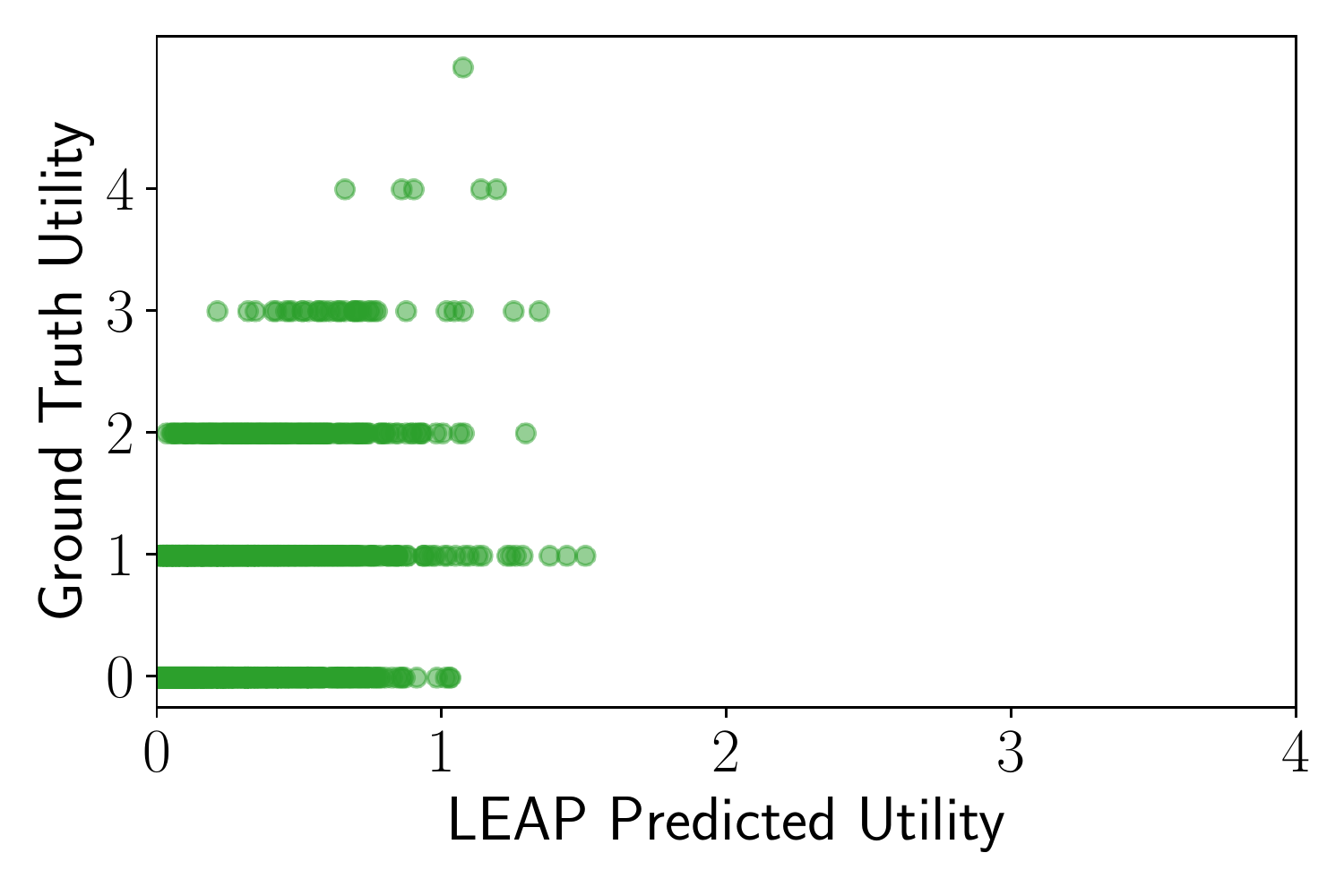}
    \includegraphics[width=1.5in]{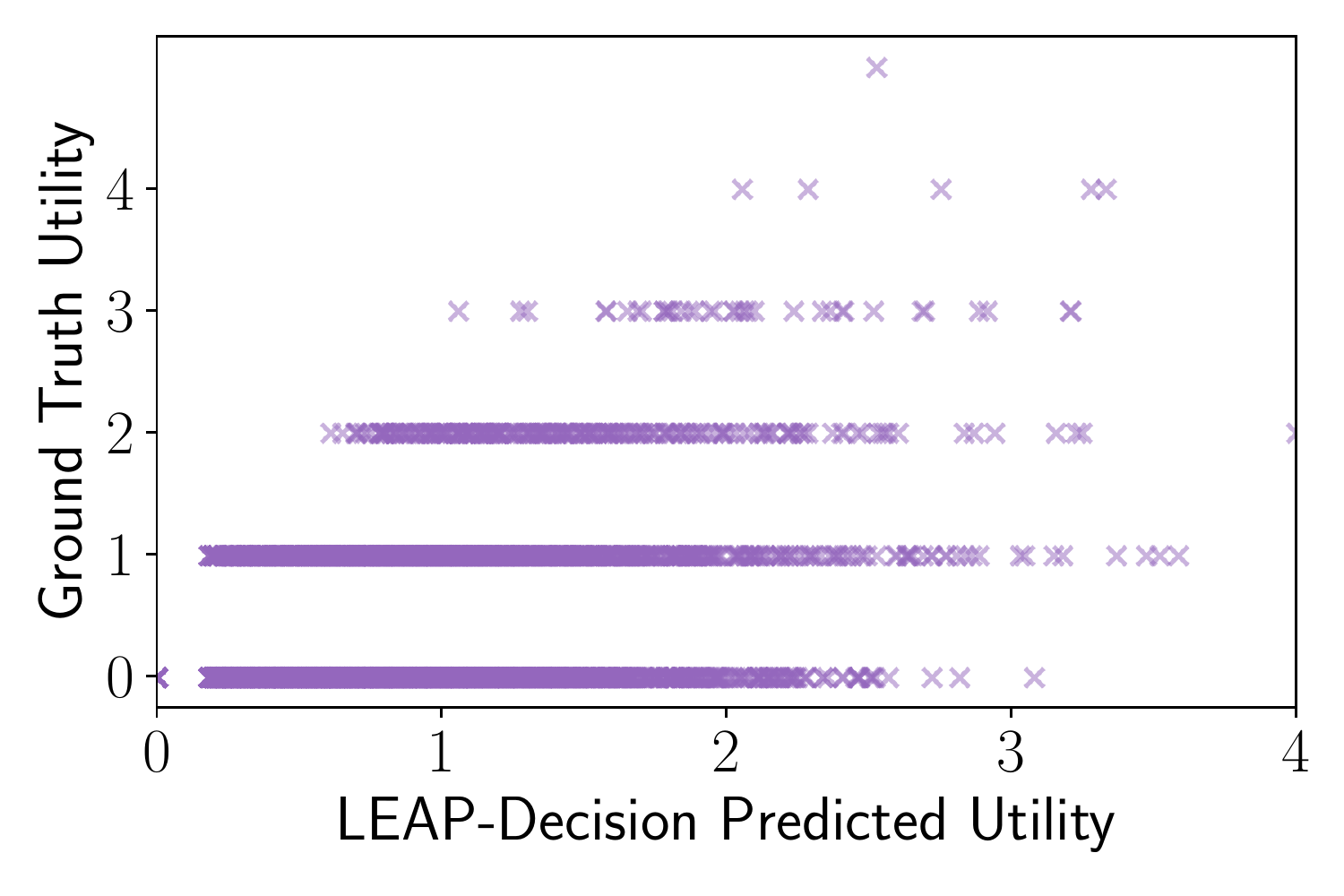}
    \caption{Results for decision focused learning problem. Top row: successful interventions and AUC for each method. Bottom row: visualizations of model predictions.}
    \label{fig:decision}
\end{figure}

\textbf{Figure \ref{fig:decision}} shows results for this task. In the top row, we see that LEAP and LEAP-Decision both outperform lw-Misses, as expected. LEAP-Decision improves the number of successful interventions by approximately 15\% compared to LEAP, demonstrating the value of tailoring the learned model to a given planning problem. LEAP-Decision actually has worse AUC than either LEAP or lw-Misses, indicating that typical measures of machine learning accuracy are not a perfect proxy for utility in decision making. To investigate what specifically distinguishes the predictions made by LEAP-Decision, the bottom row of \textbf{Figure \ref{fig:decision}} shows scatter plots of the predicted utility at each location according to LEAP and LEAP-Decision versus the true values. Visually, LEAP-Decision appears better able to distinguish the high-utility outliers which are most important to making good decisions. Quantitatively, LEAP-Decision's predictions have worse correlation with the ground truth overall (0.463, versus 0.519 for LEAP), but better correlation on locations where the true utility is strictly more than 1 (0.504 versus 0.409). Hence, decision-focused training incentivizes the model to focus on making accurate predictions specifically for locations that are likely to be good candidates for an intervention. This demonstrates the benefit of our flexible machine learning modeling approach, which can use custom-defined loss functions to automatically adapt to particular decision problems.   

\section{Discussion}

We present a framework for learning to make intervention recommendations from data generated by DAT systems applied to TB care. We develop a general approach for learning from medical adherence data that contains unobserved interventions and leverage this approach to build a model for predicting risk in multiple settings. In the real-time adherence setting, we show that our model would allow health workers to more accurately target interventions to high risk patients sooner -- catching 21\% more patients and 76\% more missed doses than the current heuristic baseline. Next, we train our model for outcome prediction, showing how adherence data can more accurately detect patients at risk of unfavorable treatment outcomes. We then derive insights that could help health workers accurately identify LCFO patients with a simple rule after just 7 days of treatment. We finally show that tailoring our LEAP model for a specific intervention via decision-focused learning can improve performance by a further 15\%. The learning approaches we present here are general and could be leveraged to study data generated by DATs as applied to any medication regimen. With the growing popularity of DAT systems for TB, HIV, Diabetes, Heart Disease, and other medications, we hope to lay the groundwork for improved patient outcomes in healthcare settings globally.

%
\begin{acks}
We thank Brandon Liu, Priyanka Ivatury, Amy Chen, and Bill Thies of Everwell for their thoughtful guidance working with the data. This research was supported by MURI Grant W911NF-18-1-0208.
\end{acks}

%
\bibliographystyle{ACM-Reference-Format}
\bibliography{5_bibliography}

\end{document}